\documentclass[11pt,a4paper]{article}
\usepackage[hyperref]{ranlp2021}
\usepackage{times}
\usepackage{multirow}
\usepackage{latexsym}

\usepackage{graphicx} 
\usepackage{color}
\newcommand\red[1]{\textcolor{red}{#1}}

\usepackage{amsmath}
\usepackage{amssymb}
\usepackage{algorithm}
\usepackage{algorithmic}
\usepackage{marvosym}

\usepackage{microtype}

\aclfinalcopy 

\title{Syntax Matters! Syntax-Controlled in Text Style Transfer}

\author{Zhiqiang Hu \\
  Singapore University of \\
  Technology and Design \\
  \texttt{hzq950419@gmail.com} \\\And
  Roy Ka-Wei Lee \\
  Singapore University of \\
  Technology and Design \\
  \texttt{roy\_lee@sutd.edu.sg} \\\And
  Charu C. Aggarwal \\
  IBM T. J. Watson \\
  Research Center \\
  \texttt{charu@us.ibm.com} \\}

\date{}

\begin{document}
\maketitle
\begin{abstract}
Existing text style transfer (TST) methods rely on style classifiers to disentangle the text's content and style attributes for text style transfer. While the style classifier plays a critical role in existing TST methods, there is no known investigation on its effect on the TST methods. In this paper, we conduct an empirical study on the limitations of the style classifiers used in existing TST methods. We demonstrate that the existing style classifiers cannot learn sentence syntax effectively and ultimately worsen existing TST models' performance. To address this issue, we propose a novel Syntax-Aware Controllable Generation (SACG) model, which includes a syntax-aware style classifier that ensures learned style latent representations effectively capture the syntax information for TST. Through extensive experiments on two popular TST tasks, we show that our proposed method significantly outperforms the state-of-the-art methods. Our case studies have also demonstrated SACG's ability to generate fluent target-style sentences that preserved the original content.
\end{abstract}

\section{Introduction}
Text Style Transfer (TST) is an increasingly popular natural language generation task that aims to change the stylistic properties (e.g., the sentiment of text) of the text while retaining its style-independent content \cite{hu2020text}. Due to the difficulty in obtaining training sentence pairs with the same content and differing styles, most existing methods are designed to perform TST in an unsupervised manner; the models only have access to non-parallel, but style-labelled sentences.

A popular TST approach is to leverage an adversarial learning autoencoder framework where a style classifier or discriminator is pre-trained to first disentangle the content and style latent representations, before using a decoder to generate the output sentence in the target style \cite{shen2017style,zhao2018adversarially,fu2018style,chen2018adversarial}. Another line of work proposed attribute-controlled generation methods where the style attribute latent vector is learned and combine with the latent representation of the text to generate output sentences in target style \cite{hu2017toward,dai2019style,zhang2018shaped}. Similar to the adversarial learning approach, the learning of the style attribute latent vector is guided using a pre-trained style classifier.


A common key component in the two aforementioned TST approaches is the usage of a style classifier. However, little is known about the effects of the style classifier on these models. For instance, is the style classifier effective in learning the style in the text? What aspects of the text style has the existing style classifier learned? Can the style classifiers distinguish text's syntax? Can the style classifier guide TST models to generate syntactically correct sentences and in the target style? This paper investigates these questions by conducting an empirical analysis of the style classifiers used in TST models. 



Extending from our empirical study, we propose the Syntax-Aware Controllable Generation (SACG)\footnote{Code implementation: https://gitlab.com/bottle\_shop/ snlg/style/sacg} model, which includes a syntax-aware style classifier that ensure that the learned style latent representations effectively capture the syntax information for TST. Through extensive experiments with two popular TST datasets and human evaluation, we demonstrated SACG's ability to outperform the state-of-the-art baselines in the TST tasks. 


\section{Related Work}
In recent years, studies on text style have attracted not only the linguist's attention but also that of many computer science researchers. Specifically, computer science researchers are investigating the Text Style Transfer (TST) task that aims to change the text's stylistic properties while retaining its style-independent content. The recent comprehensive survey \cite{hu2020text} summarizes the existing TST approaches. 


Among these approaches, a popular line of research aims to infer a latent representation for an input sentence and manipulate the generated sentence's style based on this learned latent representation. Two techniques are commonly used to learn and manipulate the text's style latent representations: (1) adversarial learning and (2) attribute controlled generation. Shen et al. \shortcite{shen2017style} leverages an adversarial training scheme where a classifier is used to evaluate if an encoder is able to generate a latent content representation devoid of style. The text content latent representation is subsequently used to generate a specific style sentence using a style-dependent decoder. Similar works have been proposed where a classifier is pretrained to enable the adversarial learning process in TST models \cite{zhao2018adversarially,fu2018style,chen2018adversarial,logeswaran2018content,yin2019utilizing,lai2019multiple,john2019disentangled}.

Hu et al. \shortcite{hu2017toward} proposed an attribute-controlled generation text style transfer model that utilized a Variational Autoencoder (VAE) \cite{kingma2013auto} to learn a sentence's latent representation $z$ and leverage a style classifier to learn a style attribute vector $s$. Subsequently, $z$ and $s$ are input into a decoder to generate a target style sentence. Similar attribute-controlled generation methods have been proposed for the TST task \cite{dai2019style,zhang2018shaped,li2019domain}.

In the aforementioned methods, pretrained style classifiers played a vital role in guiding the TST task. However, these style classifiers are often pre-trained without considering the syntax of sentences. We postulate that syntax is an important aspect of text style, especially in text formality style transfer. This paper empirically demonstrates the importance of modeling syntax in the TST task and proposes a novel syntax-aware TST method that outperforms state-of-the-art TST methods.

\section{Empirical Study}
\label{sec:empirical}
\begin{table}[t]
\small
\centering
\begin{tabular}{c|c|ccc}
\hline
\textbf{Classifier} & \textbf{Test set} & \textbf{ACC} & \textbf{F} & \textbf{I} \\
\hline
\multirow{2}*{TextCNN} & GYAFC & 88.6 & 91.3 & 86.4 \\
& Disordered & 85.3 & 84.9 & 85.5 \\
\hline
\multirow{2}*{RNN} & GYAFC & 85.6 & 84.6 & 86.4 \\
& Disordered & 82.2 & 74.8 & 87.8 \\
\hline
\multirow{2}*{Transformer} & GYAFC & 84.9 & 86.7 & 83.7 \\
& Disordered & 82.9 & 80.5 & 84.6 \\
\hline
\end{tabular}
\caption{Style classifiers performance on \textit{GYAFC} test set and corresponding \textit{Disordered} test set. \textbf{ACC} refers to the accuracy on both formal and informal sentences, \textbf{F} refers to the accuracy on formal sentences, and \textbf{I} refers to the accuracy to informal sentences.}
\label{tbl:empirical_study_results}
\end{table}


Before presenting our proposed method, we first conduct an empirical study on the style classifiers used in existing TST methods. The goal is to examine the style classifiers' ability to learn the syntax style information in a given text. 


TextCNN~\cite{kim2014convolutional}, RNN~\cite{cho2014learning}, and Transformer~\cite{vaswani2017attention} are popular style classifiers used in many TST models \cite{dai2019style,john2019disentangled,Luo19DualRL,li2019domain,zhang2018style}. In this study, we train the three style classifiers on \textit{GYAFC}~\cite{rao2018dear}, which is a popular formality transfer dataset used in many TST studies. We first train and test the classifiers using the original GYAFC training and test set. Next, we perturb the sentence structure of the text in the GYAFC test set by disordering the sentences' word order. The underlying intuition is that there should be syntax differences between formal and informal sentences, and the style classifiers should be able to learn the syntactic style information. Therefore, perturbing the test set's sentence structure should worsen classification accuracy as the syntactic information in the text is corrupted.



The empirical experiment results show that syntax plays a crucial role in text's formality. Table \ref{tbl:empirical_study_results} shows the results of our empirical experiments. We observed a small 2.9\% decrease in style classification accuracy in the \textit{disordered} test set compared to the original GYAFC test set. We further examined the style classifiers' performance in different classes. We noted that the classification accuracy for formal sentences sharply decreased as we disordered the test sentences' word order. However, such observations are not made for informal sentences; the classification accuracy remained fairly constant even when word order was disrupted in informal sentences. From the observations, we postulate that the style classifiers may have focused on the attribute words to predict the style of sentences while neglecting the syntactic information in their style predictions. Furthermore, the style classifiers may have regarded the perturbed sentences as informal ones. Nevertheless, the syntax of informal sentences should be different from the perturbed sentences. The similar classification performance on perturbed sentences demonstrated the style classifiers' ineffectiveness in capturing different formality styles' syntax information. More importantly, the style classifier's inability to learn syntax information could misguide the TST model's decoder to generate fragmented sentences, especially when transferring sentences to the informal style.

\section{Methodology}
This section proposes the Syntax-Aware Controllable Generation (SACG) model, which addresses the ineffectiveness of existing TST methods in handling sentence structure when transferring text style. We first introduce  Graph Convolutional Networks (GCNs). Subsequently, we explain how the GCNs are utilized to extract sentence structure information in our syntax-classifier and syntax-encoder, which are the two main components in our proposed SACG model. Finally, we describe the learning process of our SACG model.

\subsection{GCN and Sentence Structure Representation}
As a variant of convolutional neural networks \cite{lecun1998gradient}, graph convolutional networks (GCN) \cite{DBLP:conf/iclr/KipfW17} is designed for graph data and it has demonstrated effectiveness in modeling text data via syntactic dependency graphs \cite{marcheggiani2017encoding}. Consider a graph $\mathcal{G}=\{\mathcal{V}, \mathcal{E}\}$ where $\mathcal{V}$ (where $|\mathcal{V}|=n$ is the number of vertices in $\mathcal{G}$) is the set of graph node and $\mathcal{E}$ is the set of graph edges. Given a feature matrix $X \in \mathbb{R}^{n \times d} $, where row $x_i \in \mathbb{R}^d$ corresponds to a feature for vertex $i$, the propagation rule of a GCN is given as 

\begin{equation}
    H^{(l+1)}=\sigma(AH^{(l)}W^{(l)}),
\end{equation}

where $H^{(l)} \in \mathbb{R}^{n \times d_l}$ is the feature matrix of the $l$-th layer and $d_l$ is the number of features for each node in the $l$-th layer. $H^{(0)}=X$, $W^{(l)}$ is the weight matrix between the $l$-th and $(l+1)$-th layers, $A \in \mathbb{R}^{n \times n}$ is the adjacency matrix associated with the graph $\mathcal{G}$, and $\sigma(\cdot)$ is a non-linear activation function, such as ReLU or Leaky ReLU. In essence, a GCN takes in a feature matrix $X$ as an input and extract a latent feature matrix $H^{(L)}$ as the output, where $L$ is the number of layers in GCN.


Our goal is to extract and utilize sentence structure information to guide our SACG model to generate more plausible sentences. The syntactic relations between words in a sentence can be represented using dependency trees \cite{marcheggiani2017encoding}. 
A dependency tree can be regarded as a directed graph, and the GCNs can be used to extract the latent representation of sentence structure from the dependency trees. Previous studies have attempted to use GNCs to learn syntactic representation from dependency trees \cite{marcheggiani2017encoding,bastings2017graph}. However, many of these existing techniques are over-parameterized, especially on huge datasets. To overcome this limitation, we employ a simpler approach where an adjacency matrix incorporated with directions is used to represent a sentence's structure. Specifically, the adjacency matrix $A$ is used to represent the dependency relations of all words in the sentence. The column words are head words, and the row words are dependents. We set the element $A_{ij}$ to 1 if there is a dependency between the $i$-th word (head) and the $j$-th word (dependent). Similar to \cite{marcheggiani2017encoding}, we add a self-loop for each node in the graph, where all diagonal elements of $A$ are set to 1.


\subsection{Syntax-Aware Style Classifier}
In this subsection, we propose syntax-aware style classifier $D$ to encode the syntactic information from the dependency trees better.   

\begin{figure}[t] 
	\centering
	\setlength{\tabcolsep}{0pt} 
	\renewcommand{\arraystretch}{0} 
	\begin{tabular}{c}
		\includegraphics[scale = 0.5 ]{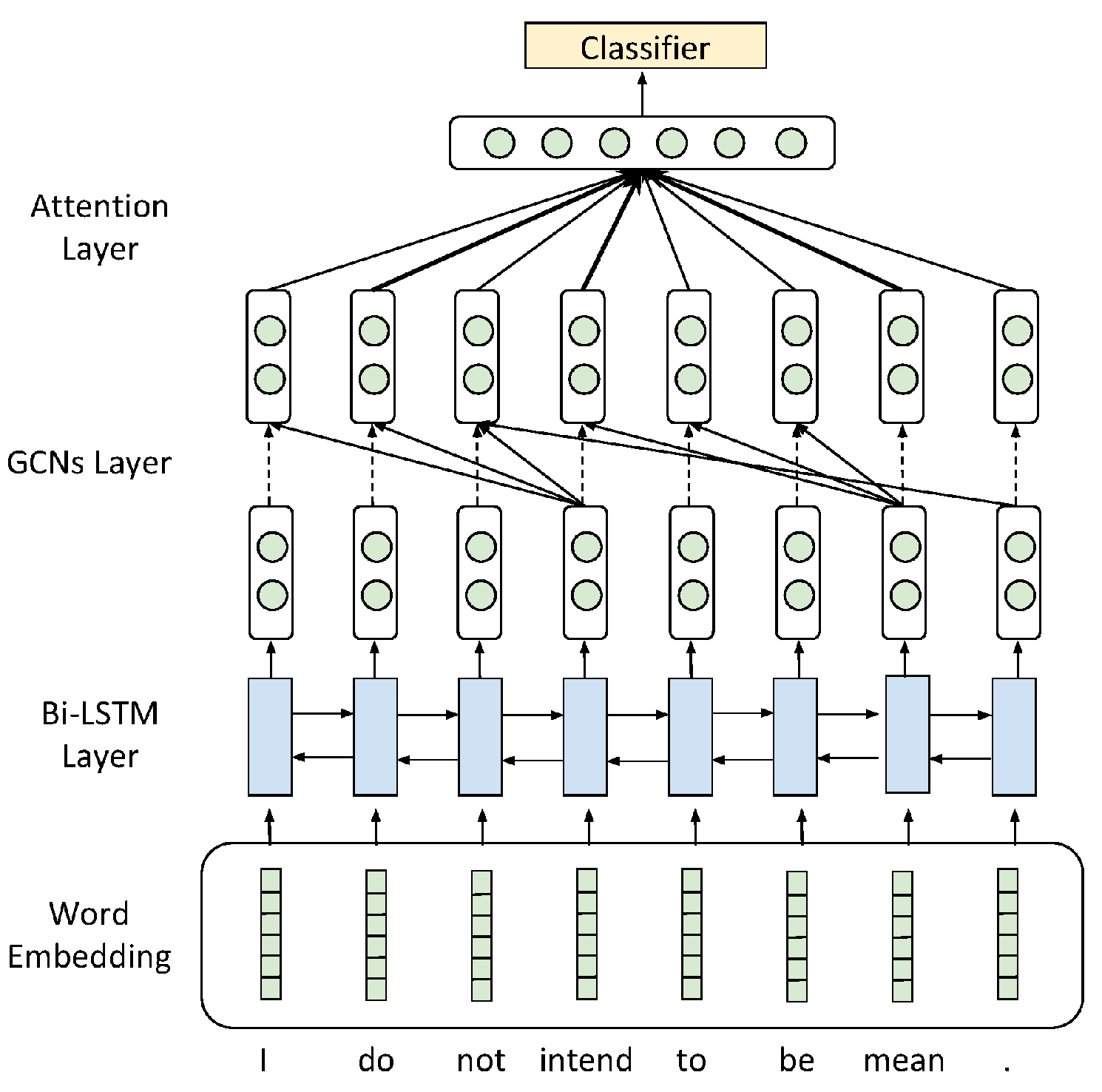}
	\end{tabular}
	\caption{Architecture of syntax-aware style classifier.}
	\label{fig:syntax-classifier}
\end{figure}

\begin{figure*}[t] 
	\centering
	\setlength{\tabcolsep}{0pt} 
	\renewcommand{\arraystretch}{0} 
	\begin{tabular}{c}
		\includegraphics[scale = 0.5 ]{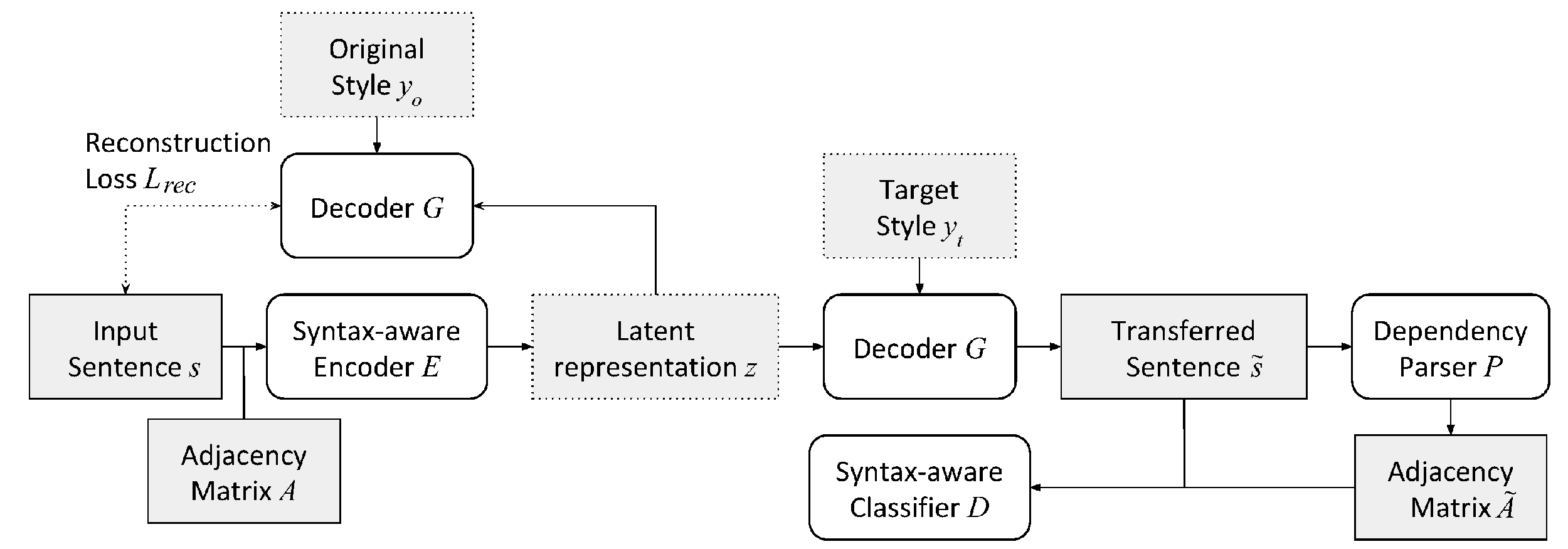}
	\end{tabular}
	\caption{Framework of the Syntax-Aware Controllable Generation (SACG) model.}
	\label{fig:SACG}
\end{figure*}

Figure \ref{fig:syntax-classifier} shows the architecture of our proposed syntax-aware style classifier. We first encode the tokens in a sentence of size $n$ as $s = \{w_1,...,w_n\}$ in the word embedding layer, where $w_i$ is the $i$-th step input of Bi-LSTM. GCN has a limitation in capturing dependencies between nodes far away from each other in the graph. Therefore, instead of performing the graph convolution on the static word embeddings, we perform the GCN operations on top of the Bi-LSTM hidden states~\cite{marcheggiani2017encoding}. As such, the GCN will only need to model the relationships for fewer hops. The Bi-LSTM states $H_{lstm}=\{h_{lstm,1},...,h_{lstm,n} \}$ serve as input $x_i=h_{lstm,i}$ to GCN, where $h_{lstm,i}$ is the concatenation of the forward and backward hidden states. We feed the hidden states into a $L$-layer GCN to obtain the hidden representations of each token, which are directly influenced by its neighbors no more than $L$ edges apart in the dependency tree. Formally, the hidden representation of node $i$ at the $(l+1)$-th layer of GCN is computed by the following equation:


\begin{equation}
    h_i^{(l+1)} = \sigma (\sum_{j=1}^nA_{ij}W^{(l)}h_j^{(l)}+b^{(l)})
\end{equation}

where $A$ is the adjacency matrix of dependency tree, $W^{(l)}$ and $b^{(l)}$ are the model parameters, and $\sigma$ is an activation function. We obtain the hidden representation $h_i^{(L)}$ of node $i$ after $L$ GCN layers.

We noted that some node representations are more informative by gathering information from syntactically related neighbors through GCN. Thus, we utilize scaled dot-product attention \cite{vaswani2017attention} and averaging to aggregate the node representations to sentence representation:

\begin{equation}
    Attention(Q,K,V) = softmax(\frac{QK^T}{\sqrt{d_k}})V
\end{equation}

where $Q,K,V$ represent queries, keys, and values, respectively, $\frac{1}{\sqrt{d_k}}$ is the scaling factor. In practice, we feed the output $H^{(L)}$ of GCN to $Q,K,V$. Finally, we obtain the style prediction by feeding the sentence representation into a fully connected neural network followed by the softmax operation.

\subsection{Syntax-aware Controllable Generation}
Figure \ref{fig:SACG} shows the framework of our proposed Syntax-Aware Controllable Generation (SACG) model. For each input sentence $s$ with attribute $y_o$ and the corresponding adjacency matrix $A$, the syntax-aware encoder $E$ encodes $s$ to a latent representation $z=E(s,A)$. $E$ is designed to extract sentence structure using the feature extractor of our proposed syntax-aware classifier. Subsequently, a decoder $G$ decodes transferred sentence $\tilde{s}=G(z,y_t)$ or input sentence $s=G(z,y_o)$ based on the attribute controlling code $y_t$ or $y_o$. We employ the Stanford neural dependency parser Stanza \cite{zhang2020biomedical} to generate the dependency tree for transferred sentences, and the corresponding adjacency matrix $\tilde{A}$. The transferred sentence $\tilde{s}$ and the corresponding adjacency matrix serve as the input of the syntax-aware classifier $D$, and the classifier will evaluate if the transferred sentence has the desired style.




We train the SACG model with classification loss $L_{cla}$ and reconstruction loss $L_{rec}$. 

\textbf{Classification Loss $L_{cla}$}: The classification loss ensures the transferred sentence is in the target style. To this end, we apply the pretrained syntax-aware classifier to guide the updates of related parameters such that the output sentence is predicted to be in the target style:

\begin{equation}
    L_{cla} = - \mathbb{E}_{(s,y_o) \sim D}[logP(y_t|G(\tilde{s}),\tilde{A})]
\end{equation}

where  $G(\tilde{s})$ denotes a soft generated sentence based on Gumbel-Softmax distribution \cite{DBLP:conf/iclr/JangGP17} and the representation of each word is defined as the weighted sum of word embeddings with the prediction probability at the current timestep. $\tilde{A}$ denotes the corresponding adjacency matrix of transferred sentence $\tilde{s}$.

\textbf{Reconstruction Loss $L_{rec}$}: The reconstruction loss attempts to preserve the original content information in the transferred sentences. Specifically, the loss function constricts the model to capture informative features to reconstruct the original sentence using the learned representations. Formally, we define $L_{rec}$ as follows:

\begin{equation}
    L_{rec} = -logP(s|z,y_o)
\end{equation}

Where $z$ denotes the hidden representation extracted by our syntax-aware encoder, and $y_o$ denotes the original style of input sentence $s$.

\textbf{Putting them together}, the final joint training loss $L$ is as follows:



\begin{equation}
    L = L_{rec} + \lambda L_{cla}
\end{equation}

Where $\lambda$ is a balancing hyper-parameter to ensure that the transferred sentence has the target style while preserving the original content. 

\section{Experiments}


\subsection{Experiment Setting}

\textbf{Datasets.} We evaluate our model on two popular style transfer tasks: (1) Sentiment transfer, and (2) formality transfer. The representative Yelp \footnote{https://github.com/shentianxiao/language-style-transfer} restaurant reviews dataset \cite{shen2017style} is selected for the sentiment transfer task. Following the same data preprocessing steps proposed in \cite{shen2017style}, reviews with a rating above 3 are considered positive, and those below 3 are negative. We adopt the same train, development, and test split as \cite{shen2017style}. Rao et al. \shortcite{rao2018dear} released the GYAFC \footnote{https://github.com/raosudha89/ GYAFC-corpus} (Grammarly’s Yahoo Answers Formality Corpus) dataset to facilitate the formality style transfer task. We adopt the \textit{Family\&Relationship} (F\&R) domain data for our experiments. Although it is a parallel dataset, the alignments are only used for evaluation and not for model construction. Table \ref{tbl:dataset_statistics} shows the training, validation, and test splits of the Yelp and GYAFC datasets used in our experiments.

\begin{table}
\small
\centering
\begin{tabular}{|c|c|c|c|c|}
\hline
Dataset &  Attributes & Train & Dev & Test \\
\hline
\multirow{2}*{Yelp}  & Positive & 267K & 38K & 76K \\
 & Negative & 176K & 25K & 50K \\
\hline
\multirow{2}*{GYAFC}  & Informal & 51K & 2.7K & 1.3K \\
 & Formal & 51K & 2.2K & 1K \\
\hline
\end{tabular}
\caption{Dataset statistics for Yelp and GYAFC.}
\label{tbl:dataset_statistics}
\end{table}

\textbf{Baselines}. We benchkmark SACG against 12 state-of-the-art TST models:\textit{ARAE} \cite{zhao2018adversarially}, \textit{DualRL} \cite{Luo19DualRL}, \textit{DAST, DAST-C} \cite{li2019domain}, \textit{PFST} \cite{He2020A},   \textit{DRLST} \cite{john2019disentangled},  \textit{DeleteOnly, Template, Del\&Retri} \cite{li2018delete}, \textit{DIRR} \cite{liu-etal-2021-learning}, and \textit{HPAY} \cite{kim-sohn-2020-positive}. 


\textbf{Training.} The experiments were performed on an Ubuntu 18.04.4 LTS system with 24 cores, 128 GB RAM, and Nvidia RTX 2080Ti. The word embeddings of 300 dimensions are learned from scratch. We use a single Bi-LSTM layer followed by 2 GCN layers. The hidden dimension of the latent representation $z$ is set to 500, and the learnable vectors with 200 dimensions represent the style labels. The decoder is initialized by a concatenation of the latent representation $z$ and attribute controlling code $y$. The syntax-aware style classifier is pretrained for evaluation and guiding the decoder's generation. After pretraining, the parameters of the classifier are fixed. We use the Gumbel-softmax to back-propagate the loss through discrete tokens from the classifier to the encoder-decoder model \cite{JanGuPoo17}. We empirically set the learning rate to $1\times 10^{-5}$ and the balancing parameter $\lambda$ to 1. 

\begin{table*}[t]
\small
\centering
\begin{tabular}{cccccccc}
\hline
\textbf{Model} & \textbf{ACC(\%)} & \textbf{BLEU} & \textbf{CS} & \textbf{WO} & \textbf{PPL} & \textbf{G-Score} \\
\hline
ARAE \cite{zhao2018adversarially}      & 76.2 & 2.2 & 0.903 & 0.042 & 35 &0.71 \\
DeleteOnly \cite{li2018delete} & 18.7 & 16.2 & 0.945 & 0.431 & 74 &1.11 \\
Template \cite{li2018delete}  & 44.7 & 19.0 & 0.943 & 0.509 & 102 &1.32 \\
Del\&Retri \cite{li2018delete} & 50.7 & 11.8 & 0.934 & 0.345 & 74 &1.21 \\
DualRL \cite{Luo19DualRL}    & 59.8 &  18.8 & 0.944 & 0.447 & 266 &1.12 \\
DAST \cite{li2019domain}      & 78.3 & 14.3 & 0.934 & 0.350 & 352 &1.01 \\
DAST-C \cite{li2019domain}    & 79.2 & 13.8 & 0.927 & 0.328 & 363 &0.98 \\
DRLST \cite{john2019disentangled}     & 49.8 & 2.7 & 0.909 & 0.342 & \textbf{31} &1.06 \\

PFST \cite{He2020A}      & 48.3 & 16.5 & 0.940 & 0.393 & 116 &1.25 \\
HPAY \cite{kim-sohn-2020-positive} & 43.1 & 10.4 & 0.942 & 0.418 & 92 & 1.17 \\
DIRR \cite{liu-etal-2021-learning} & 71.8 & 18.2 & 0.942 & 0.451 & 145 & 1.28 \\
SACG (ours) & \textbf{84.1} & \textbf{21.1} & \textbf{0.962} & \textbf{0.591} & 73 &\textbf{1.69} \\
\hline
Human0     & \textbf{84.6} & 24.6 & \textbf{0.942} & \textbf{0.393} & \textbf{24} &\textbf{2.00} \\
Human1     & 83.8 & 24.3 & 0.931 & 0.342 & 27 &1.89 \\
Human2     & 83.6 & 24.6 & 0.932 & 0.354 & 27 &1.91 \\
Human3     & 82.1 & \textbf{24.7} & 0.931 & 0.354 & 27 &1.90 \\
\hline
\end{tabular}
\caption{Performance of models on GYAFC dataset (Formality Transfer Task).}
\label{tbl:GYAFC_results}
\end{table*}

\begin{table*}[t]
\small
\centering
\begin{tabular}{ccccccc}
\hline
\textbf{Model} & \textbf{ACC(\%)} & \textbf{\textit{self}-BLEU} & \textbf{CS} & \textbf{WO} & \textbf{PPL} & \textbf{G-Score} \\
\hline
ARAE \cite{zhao2018adversarially}      & 83.2 & 18.0 & 0.874 & 0.270 & 79 &1.35 \\
DeleteOnly \cite{li2018delete} & 84.2 & 28.7 & 0.893 & 0.501 & 130 & 1.53 \\
Template \cite{li2018delete}  & 78.2 & 48.1 & 0.850 & 0.603 & 250 & 1.50  \\
Del\&Retri \cite{li2018delete} & 88.1 & 30   & 0.897 & 0.464 & 88  & 1.66 \\
DualRL \cite{Luo19DualRL}    & 79.0 & \textbf{58.3}  & 0.970 & \textbf{0.801} & 117 &1.98 \\
DAST \cite{li2019domain}      & 90.7 & 49.7  & 0.961 & 0.705 & 181 &1.76 \\
DAST-C \cite{li2019domain}    & 93.6 & 41.2  & 0.933 & 0.560 & 274 &1.49 \\

DRLST \cite{john2019disentangled}     & 91.2 & 7.6  & 0.904 & 0.484 & \textbf{65}  &1.36  \\
PFST \cite{He2020A}      & 85.3 & 41.7 & 0.902 & 0.527 & 94 &1.78 \\
HPAY \cite{kim-sohn-2020-positive} & 86.5 & 31.2 & 0.886 & 0.450 & 85 & 1.66 \\
DIRR \cite{liu-etal-2021-learning} & \textbf{94.2} & 52.6 & 0.957 & 0.715 & 292 &1.63  \\
\hline
SACG (ours) & 93.0 & 57.7 & \textbf{0.971} & 0.778 & 74 &\textbf{2.23} \\
\hline
\end{tabular}
\caption{Performance of models on Yelp dataset (Sentiment Transfer Task). }
\label{tbl:yelp_results}
\end{table*}

\subsection{Automatic Evaluation}

We evaluate the proposed model and baselines on three criteria commonly used in TST studies: \textit{transfer strength}, \textit{content preservation}, and \textit{fluency}.

\textbf{Transfer strength.} A TST model's transfer strength or its ability to transfer text style is commonly measured using \textit{style transfer accuracy} \cite{hu2020text}. A syntax-aware style classifier is first pre-trained to predict the style label of the input sentence. The classifier is subsequently used to approximate the style transfer accuracy of the sentences' transferred style by considering the target style as the ground truth.  

\textbf{Content preservation.} To quantitatively measure the amount of original content preserved after the style transfer operation, we employed four metrics used in previous work \cite{fu2018style,john2019disentangled,He2020A}:

\begin{itemize}
    \item \textit{BLEU}: The BLEU score \cite{papineni2002bleu} is used to compare  the style transferred sentences with the human references provided in the GYAFC dataset.
    \item \textit{self-BLEU}: The \textit{self}-BLEU score is adopted by comparing the style transferred sentence with its original sentence. This metric is used when human reference is not available. 
    \item \textit{Cosine Similarity}: Fu et al. \shortcite{fu2018style} calculated the cosine similarity between original sentence embedding and transferred sentence embedding. The two sentences' embeddings should be close to preserve the semantics of the transferred sentences.
    \item \textit{Word Overlap}: Vineet et al. \cite{john2019disentangled} employed a simple metric that counts the unigram word overlap rate of the original and style transferred sentences.
    
\end{itemize}

\textbf{Fluency.}  Generating fluent sentences is a common goal for most natural language generation models. GPT-2 \cite{radford2019language} is a large-scale transformer-based language model that is pretrained on large text corpus. We fine-tuned GPT-2 on the GYAFC and Yelp datasets and use the model to measure the perplexity (PPL) of transferred sentences. The sentences with smaller PPL scores are considered more fluent. 

\textbf{Geometric Mean (G-Score)}: We compute the geometric mean of \textit{ACC}, \textit{\textit{self}-BLEU}, \textit{BLEU}, \textit{CS}, \textit{WO} and \textit{1/PPL}.
Notably, we take the inverse of the calculated perplexity score because a smaller PPL score corresponds to better fluency.

\subsubsection{Automatic Experiment Results}

Table \ref{tbl:GYAFC_results} shows the performance of the proposed SACG model and baselines on the formality transfer task. SACG has achieved the best G-Score, outperforming the state-of-the-art baselines. Nevertheless, we noted that none of the TST models could score well on all evaluation metrics. Many of the baselines can only perform well on transfer strength or content preservation, but not on both evaluation criteria. SACG has outperformed the baselines in G-Score, and achieve 84.1\% transfer accuracy and 21.1 average BLEU score. The GYAFC dataset also provided the performances of four human references performing the formality transfer task on the test set. The BLEU score of each human reference is calculated with the other three human references. Interestingly, we observe that SACG's performance on the three TST evaluation criteria is comparable and close to human references' performance.



Similar results were observed for the sentiment transfer task. Table \ref{tbl:yelp_results} shows the performance of the proposed SACG model and baselines on the Yelp dataset. We computed the  self-BLEU scores as no human references are provided for the Yelp test set. Similarly, SACG outperformed the baselines in G-score. We observe that the average style transfer accuracy in Yelp is 86.3\%, which is significantly higher than GYAFC's average score of 66.0\%. The difference in the average style transfer accuracy highlights the challenge of the formality transfer task. We also noted that most models performed better in this task compared to the formality transfer task. Nevertheless, the trade-off phenomenon between transfer strength and content preservation is still observed in the sentiment transfer task. 

\begin{table}[t]
\small
\centering
\begin{tabular}{cccc}
\hline
\textbf{Model} & \textbf{Style(\%)} & \textbf{Content} & \textbf{Fluency} \\
\hline
DualRL & 28.5 & 4.09 & 4.52 \\
DAST & 27.5 & 3.22 & 3.68 \\
PFST & 24.0 & 3.91 & 4.54 \\
Del\&Retri & 25.5 & 2.61 & 3.23 \\
SACG & \textbf{44.5} & \textbf{4.39} & \textbf{5.07} \\
\hline
\end{tabular}
\caption{Human evaluation results on GYAFC dataset.}
\label{tbl:human_evaluation}
\end{table}

\subsection{Human Evaluation}

To further evaluate SACG's performance in generating syntactically correct sentences in target style, we conducted a human-based evaluation study. Specifically, we first randomly sampled 200 sentences from the GYAFC dataset. Next, we perform text style transfer for the sampled sentences using SAGC and four competitive baselines. Finally, we recruited two linguistics researchers (i.e., participants) to evaluate the style-transferred sentences generated by the TST models. The participants are asked to evaluate the generated sentences on the three criteria discussed in the earlier section. Specifically, for \textit{Transfer Strength}, participants are asked to indicate if the generated sentences are in the target style (i.e., a binary true/false indicator). For \textit{Content Presentation}, the participants are asked to rate the amount of content preserved in the generated sentences using a 6-point Likert scale. 1: no content presented, and 6: all content are preserved. Similarly, for \textit{Fluency}, the participants are asked to rate fluency in the generated sentences using a 6-point Likert scale. 1: too many grammatical errors, and 6: perfect and fluent sentence.

\begin{table}[t]
\small
\centering
\begin{tabular}{cccc}
\hline
\textbf{Model} & \textbf{TED} & \textbf{Model} & \textbf{TED} \\
\hline
DRLST & 19.2 & DeleteOnly & 18.2  \\
ARAE & 18.1 & Template & 17.9 \\
DualRL & 15.2 & Del\&Retri & 21.0 \\
DAST & 16.6 & HPAY & 18.4 \\
PFST & 15.5 & DIRR & 15.5 \\
DAST-C & 16.9 & SACG (ours) & \textbf{13.2} \\
\hline
\end{tabular}
\caption{Average Tree Edit Distance (TED) of constituency tree between TST model generated sentences and 4 human references in GYAFC. }
\label{tbl:efficacy_study}
\end{table}


To minimize biases, we do not display the models' names and we shuffled the order of the models when displaying their generated sentence. Therefore, the participants do not know which model generates a particular sentence.

\subsubsection{Human Experiment Results}

Table \ref{tbl:human_evaluation} shows the human evaluation results. For the transfer style, we compute the models' style transfer accuracy using the binary feedback from the participants. We compute the models' average 6-point Likert scores for content preservation and fluency criteria. SACG is observed to outperform the baselines in all three criteria. SACG is also rated to generate more syntactically sound and fluent sentence compared to the baselines. To check for participant bias, we compute the inter-annotator agreement between the participants. The Cohen's kappa coefficients on style transfer strength, content preservation, and fluency are 0.54, 0.76, and 0.72, respectively. The participants have substantially high agreement on the content presentation and fluency. However, the participants' agreement for style transfer strength is moderate as text formality is subjective, and the participants are only asked to perform binary indication.

\subsection{Syntax Evaluation}

\begin{table*}[t]
\small
\centering
\begin{tabular}{ccccccc}
\hline
\textbf{Model} & \textbf{ACC(\%)} & \textbf{\textit{self}-BLEU} & \textbf{BLEU} & \textbf{CS} & \textbf{WO} & \textbf{PPL} \\
\hline
\multicolumn{7}{c}{GYAFC} \\
\hline
SACG                      & 84.1 & - & 21.1 & 0.962 & 0.591 & 73 \\
SACG w/o Syntax-aware Encoder    & 83.8 & - & 20.3 & 0.957 & 0.544 & 83  \\
SACG w/o Syntax-aware Encoder \& Classifier  & 78.7 & - & 15.6 & 0.943 & 0.446 & 223 \\

\hline
\multicolumn{7}{c}{Yelp} \\
\hline
SACG                      & 93.0 & 57.7 & - & 0.971 & 0.778 & 74 \\
SACG w/o Syntax-aware Encoder    & 92.6 & 56.4 & - & 0.964 & 0.720 & 85  \\
SACG w/o Syntax-aware Encoder \& Classifier & 89.3 & 49.1 & - & 0.943 & 0.697 & 230 \\ \hline
\end{tabular}
\caption{Results of ablation study.}
\label{tbl:ablation_study}
\end{table*}

\begin{table*}[t]
\small
\centering
\begin{tabular}{c|c|c}
\hline
& From formal to informal(GYAFC) & From positive to negative (Yelp) \\
\hline
Source & also , i dislike it when my father is unhappy . & We will definitely come back here! \\
\hline
DualRL & also \red{i thrilled}... & We will not come back here!\\
DAST & also, \red{i r it} when my father \red{is men}! & We will \red{normally joke} back here? \\
PFST & so i miss it when my father is 18. & We will not come back here again. \\
\hline
SACG (ours) & i also hate it when my father is unhappy !! & We will not come back here! \\
\hline
\end{tabular}
\caption{Example outputs on the GYAFC and Yelp datasets. Grammatical errors are \red{colored}.}
\label{tbl:case_study}
\end{table*}

As human references are available in the GYAFC dataset, we compare the syntax of the sentences generated by the TST models with the human references. Specifically, we compute the constituency tree edit distance (TED) to measure the syntactic similarity between generated sentences and human references. The intuition is that the TST model that could generate sentences with similar syntactic structure as the human references would likely have learned the syntactic information associated with the text formality style. To compute the constituency TED, we parse the sentences using Stanford CoreNLP and compute the TED between constituency parsing trees. 

Table~\ref{tbl:efficacy_study} shows the syntax evaluation results. We noted that SACG outperformed the baseline in generating sentences that are syntactically similar to human references. This superior performance in both the formality transfer task and syntax evaluation suggests that SACG is able to learn the syntax information of formal and informal text to perform better text formality transfer.

\subsection{Ablation Study}
We also conducted an ablation study to further examine the importance of syntax-aware classifier and encoder in the SACG model. Table \ref{tbl:ablation_study} shows the results of our ablation study. In the ``w/o syntax-aware encoder'' setting, we replace the syntax-aware encoder with a one-layer GRU \cite{cho2014learning}. We noted a small decrease in performance for both formality transfer and sentiment transfer tasks when the encoder is replaced. In the ``w/o syntax-aware encoder \& classifier'' setting, we further replace the  syntax-aware classifier with a TextCNN \cite{kim2014convolutional} classifier. Interestingly, we observe a sharp decrease in performance for both formality transfer and sentiment transfer tasks. In particular, the absence of the syntax-aware encoder and classifier greatly worsens the fluency of the sentences. Our ablation study noted that the syntax-aware encoder and classifier play vital roles in ensuring SACG generates fluent target-style sentences that preserve the original content.  

\subsection{Case Study}

We conduct some case studies by presenting randomly sampled examples and the corresponding style transferred output of SACG and the top three baselines ranked by G-Score. Table \ref{tbl:case_study} shows the example outputs on the GYAFC and Yelp datasets. For the Yelp dataset, we observe that DualRL, PFST, and SACG are able to transfer the sentiment of the source sentence correctly. The generated sentences are also fluent and have preserved the original content (i.e., going back to a venue). The formality transfer task is observed to be more challenging, as we noted that most of the baselines could not generate acceptable output sentences. The baselines have generated output sentences with grammatical errors, making it harder to judge if the style has been successfully transferred. Albeit the difficulty of the task, SACG is able to generate a fluent sentence that preserved the original content. 

\section{Ethical Considerations}

TST algorithms have many real-world applications. For example, these algorithms can improve target marketing messages' persuasiveness and integrate into writing tools to improve users' writing style. However, TST algorithms inherently run the risk of being misused for document forgery, impersonation, and sock-puppeting. To mitigate these risks, we will add access control to our code repository, and we would share our codes after the requester has acknowledged our ethical disclaimer.


\section{Conclusion}
In this paper, we empirically examined the style classifier used in existing TST models and demonstrated that the existing style classifier could not learn the text syntax effectively. We proposed SACG, a novel deep generative framework that considers syntax when learning style latent representation. We conducted extensive experiments on two benchmark datasets and benchmarked SACG against competitive TST models. The automatic and human-based evaluation experiment results showed that SACG outperforms state-of-the-art methods. Our case studies also demonstrated that SACG is able to generate fluent target-style sentences that preserved the original content. For future work, we will continue to explore other methods to improve the structural representations of text and incorporate them to perform better TST.

\section*{Acknowledgement}
This research is supported by Living Sky Technologies Ltd, Canada under its research exploratory funding initiatives. Any opinions, findings and conclusions or recommendations expressed in this material are those of the author(s) and do not reflect the views of Living Sky Technologies Ltd, Canada.

\bibliographystyle{acl_natbib}
\bibliography{anthology,ranlp2021}


\end{document}